\documentclass{article}

\PassOptionsToPackage{table}{xcolor}
\usepackage{PRIMEarxiv}

\usepackage[utf8]{inputenc}
\usepackage[T1]{fontenc}
\usepackage{hyperref}
\usepackage{url}
\usepackage{booktabs}
\usepackage{amsfonts}
\usepackage{nicefrac}
\usepackage{microtype}
\usepackage{fancyhdr}
\usepackage{graphicx}
\graphicspath{{../images/}}

\usepackage{amsmath,amssymb}
\usepackage{algorithmic}
\usepackage{algorithm}
\usepackage{xcolor}
\usepackage{multirow}
\usepackage{bm}
\usepackage{bbm}
\usepackage{mathtools}
\usepackage{amsthm}
\usepackage{enumitem}

\newcommand{\Ln}{\mathbb{L}^{n}_{\kappa}}
\newcommand{\dhyp}{d_{\mathbb{L}}}
\newcommand{\linner}[2]{\langle #1, #2 \rangle_{\mathcal{L}}}
\newcommand{\einner}[2]{\langle #1, #2 \rangle_{\mathcal{E}}}
\newcommand{\IoA}{\mathrm{IoA}}

\DeclareMathOperator{\arccosh}{arccosh}

\title{HyperVis: Continuous Latent Visual Relational Graphs on the Lorentz Hyperboloid \\for Compositional Reasoning
\thanks{\textit{\underline{Citation}}:
\textbf{Preprint}}
}

\author{
  Moshiur Farazi$^{1}$\thanks{Corresponding author: \texttt{moshiur.farazi@udst.edu.qa}} \quad
  Sameera Ramasinghe$^{2}$ \quad
  Mahbub Ahmed Turza$^{3}$ \quad
  Shafin Rahman$^{3}$ \\[6pt]
  $^{1}$Data Science and AI, University of Doha for Science and Technology, Qatar \\
  $^{2}$Pluralis Research, Australia \\
  $^{3}$Department of Electrical and Computer Engineering, North South University, Bangladesh \\
}

\begin{document}
\maketitle

\begin{abstract}
Vision-Language Models (VLMs) struggle with compositional reasoning that requires understanding inter-object relationships. A natural remedy is to inject explicit scene graph triplets $\langle s, p, o \rangle$ from an off-the-shelf scene graph generator (SGG), but we show this backfires: discrete text labels collide with the continuous visual modality, degrading GQA accuracy from 60.38\% to 58.86\%. We propose \textbf{HyperVis}, which bypasses the SGG semantic bottleneck entirely. From $N$ class-agnostic region proposals, we compute a dense $O(N^2)$ visual relation tensor via spatially-biased cross-attention, project it onto a Lorentz hyperboloid, and enforce hierarchy through spatial physics, namely IoA-driven entailment cones and exterior-angle repulsion. We discover that HyperVis contributes in two complementary ways:   (1) as a \emph{training-time regularizer}, the hyperbolic relational losses
  shape LoRA representations that improve generative VQA (GQA 61.03\% vs.\
  57.21\% for LoRA fine-tuning without relational losses, recovering and
  surpassing the baseline); and (2) as an \emph{inference-time relational encoder}, hyperbolic prefix tokens boost discriminative compositional scoring (SugarCrepe 79.94\%, $+$6.25pp over baseline). The learned curvature stabilises at $\kappa{=}4.0$, an order of magnitude above prior hyperbolic VLMs where $\kappa$ typically collapses toward zero, indicating that continuous visual features genuinely require the exponential volume of strongly curved space. A controlled Euclidean ablation confirms this decomposition: the relational pipeline regularises LoRA comparably in flat space (GQA 60.81\%), but the compositionality gain is specifically hyperbolic (SugarCrepe $+$4.58pp over Euclidean), with entailment loss ${\sim}6{\times}$ higher in Euclidean training. Codes are available at \url{TBA}.
\end{abstract}


% =====================================================
\section{Introduction}
\label{sec:intro}
% =====================================================

Humans interpret visual scenes not merely by recognising individual objects, but by reasoning about the rich web of relationships that connect them. Determining whether a person is \texttt{riding} a horse versus \texttt{standing beside} it requires understanding both spatial configuration and functional interaction, cues that are critical for visual question answering~\cite{goyal2017vqav2,hudson2019gqa}, compositional retrieval~\cite{thrush2022winoground}, and grounded captioning.

  Foundation VLMs such as CLIP~\cite{radford2021clip},
  BLIP-2~\cite{li2023blip2}, and LLaVA~\cite{liu2024llava,liu2024llava15}
  achieve impressive performance across many multi-modal benchmarks. While
  generative VLMs with early fusion (BLIP-2, LLaVA) handle compositionality
  better than dual-encoder models like CLIP, they still exhibit systematic
  weaknesses on benchmarks that specifically probe relational and compositional
  understanding~\cite{yuksekgonul2023aro,thrush2022winoground,hsieh2024sugarcrepe}.

  A seemingly natural fix is to attach an off-the-shelf scene graph generator
  (SGG) and feed its $\langle s, p, o \rangle$ triplets into the VLM as
  additional textual
  context~\cite{xiong2022savqa,wang2024llavassg,mitra2024ccot}. We began this
  work along precisely this path. Our experiments revealed that
  \textit{injecting SGG text labels into a foundation VLM degrades, rather than
  improves, downstream accuracy} (GQA: 58.86\% vs.\ 60.38\% baseline). We
  identify two practical limitations of this approach: (i)~it depends on an
  external SGG model whose predicate vocabulary is fixed and whose errors
  propagate as categorically wrong tokens (e.g.\ \texttt{on} vs.\ \texttt{next
  to}), and (ii)~discrete predicate labels discard the rich spatial--visual cues
   (relative pose, occlusion, pixel-level interaction) that distinguish visually
   similar but semantically distinct relationships.

We therefore pivot to a fundamentally different formulation. We argue that the right level of abstraction for relational reasoning in a foundation VLM is neither a global pooled feature nor a discrete predicate label, but a \textbf{continuous latent visual relational graph} computed directly from region features. From a small set of class-agnostic region proposals, we form a dense $O(N^2)$ tensor of   \textit{visual} relations via spatially-biased cross-attention (multi-head
self-attention over region features with learned spatial geometry biases), with no text labels, no predicate vocabulary, and no external SGG model. We then embed this graph on the \textbf{Lorentz hyperboloid} and shape its geometry using \textit{purely geometric} signals:   if region $A$ is spatially contained inside region $B$ (Intersection-over-Area $\IoA(A,B) > \tau_{\text{in}}$), embedding $A$ is forced into the entailment cone of $B$; if two regions do not overlap ($\IoA < \tau_{\text{out}}$), their embeddings are pushed apart via the exterior angle. We use hard thresholds ($\tau_{\text{in}}{=}0.8$, $\tau_{\text{out}}{=}0.05$) rather than soft weighting because the entailment cone constraint is inherently binary, a point is either inside a cone or not, and the wide dead zone ($0.05 < \IoA < 0.8$) leaves ambiguous pairs unsupervised, allowing the model to place them freely on the manifold without forcing a geometric commitment. We show that this hierarchy can be predefined from spatial geometry rather than learned from semantic labels, removing dependence on predicate taxonomies entirely. A hyperbolic Top-$K$ gate finally selects the most salient relations and injects them as visual prefix tokens into LLaVA-1.5.

We refer to this framework as \textbf{HyperVis}. An intriguing outcome of this design is that the learned curvature parameter stabilises at $\kappa{=}4.0$, an order of magnitude larger than reported in prior hyperbolic   VLMs~\cite{desai2023meru,ramasinghe2024accept}, where $\kappa$ typically collapses toward $0$, effectively converging to a flat Euclidean space (the curvature bottleneck). Continuous visual features overlap heavily in pixel space, and we hypothesise, and verify experimentally, that the manifold needs the \textit{exponential volume} of a strongly curved hyperbolic space to separate distinct relational objects without breaking spatial entailment cones.
A key empirical finding is that HyperVis contributes in \textit{two distinct and complementary} ways. First, training with hyperbolic relational losses acts as a \textbf{structural regularizer} for the LoRA adapters: LoRA fine-tuning on GQA \textit{without} relational losses degrades GQA accuracy to 57.21\% ($-$3.17pp below the 60.38\% baseline), while training \textit{with} HyperVis's angle and entailment losses recovers to 61.03\% ($+$0.65pp above baseline), even when prefix tokens are dropped at inference. Second, retaining the hyperbolic prefix tokens at inference time provides an \textbf{explicit relational encoder} for discriminative compositional scoring, boosting SugarCrepe from 73.69\% to 79.94\% ($+$6.25pp). Our contributions are:
  \begin{itemize}[noitemsep]
  \item We show that continuous visual latents can better encode hyperbolic geometry, and empirically demonstrate that textual SGG triplet injection degrades VLM accuracy (GQA: 58.86\% vs.\ 60.38\% baseline).
  \item We propose a fully visual relation tensor computed via spatially-biased cross-attention over class-agnostic regions, embedded on the Lorentz hyperboloid with a \textbf{geometric IoA-driven hierarchy}: spatial containment yields entailment cones, spatial separation yields angular repulsion. A hyperbolic Top-$K$ gate selects the most salient relations as visual prefix tokens for LLaVA-1.5.
  \item We demonstrate a \textbf{dual contribution}: hyperbolic losses regularize LoRA for generative VQA (GQA 61.03\%, $+$3.82pp over LoRA-only), while prefix tokens boost compositional scoring (SugarCrepe 79.94\%, $+$6.25pp over baseline). A controlled Euclidean ablation confirms that the compositionality gain is specifically hyperbolic ($+$4.58pp over Euclidean, entailment loss ${\sim}6{\times}$ higher in flat space), while the learned curvature $\kappa{=}4.0$ challenges the conventional curvature-bottleneck narrative.
  \end{itemize}

% =====================================================
\section{Related Work}
\label{sec:related}
% =====================================================

\noindent\textbf{Vision-Language Models.}
Early VL methods learned separate representations for images and text, combining them through concatenation or bilinear pooling~\cite{vinyals2015show,fukui2016multimodal}. Transformer-based models such as VL-BERT~\cite{su2019vlbert}, ViLBERT~\cite{lu2019vilbert}, and UNITER~\cite{chen2020uniter} aligned object proposals with contextual word embeddings. The current paradigm centres on contrastive pre-training (CLIP~\cite{radford2021clip}), generative bootstrapping (BLIP-2~\cite{li2023blip2}), and visual instruction tuning (LLaVA~\cite{liu2024llava15}, Qwen2-VL~\cite{wang2024qwen2vl}, InternVL~\cite{chen2024internvl}). Despite strong overall performance, these models remain fragile on compositional benchmarks~\cite{yuksekgonul2023aro,thrush2022winoground}.

\noindent\textbf{Explicit (textual) scene graphs in VLMs.}
A long line of work attaches an SGG model~\cite{xu2017scenegraph,zhang2019graphical,yang2022panoptic,li2024pixelstographs} to a VLM and feeds the resulting predicate triplets as additional context. Examples include adaptive scene graph tokens for CLIP/BLIP-2~\cite{herzig2023structured}, the Scene Graph Expression module in LLaVA-SG~\cite{wang2024llavassg}, chain-of-thought scene graph prompts~\cite{mitra2024ccot}, end-to-end triplet detectors~\cite{im2024egtr,li2024predicate}, and the structured triplet encoder of SA-VQA~\cite{xiong2022savqa}. All of these treat predicates as a closed (or open) \textit{vocabulary} and route relational signal through \textit{discrete text}, inheriting the SGG model's predicate noise.

\noindent\textbf{Continuous latent graphs.}
A complementary strand sidesteps the SGG vocabulary entirely. Visual relationship attention~\cite{krishna2017visualgenome}, structured prediction over region proposals~\cite{xu2017scenegraph,zellers2018motifs}, and Q-Former-style latent queries~\cite{li2023blip2} all model relations as continuous tensors learned end-to-end. Our work is a hyperbolic instantiation of this view: we never materialise predicate labels, treating the $O(N^2)$ pairwise relation tensor as the primary representation and shaping it geometrically rather than semantically.

\noindent\textbf{Hyperbolic VLMs and the modality gap.}
Hyperbolic spaces embed hierarchical structures with lower distortion than Euclidean spaces~\cite{nickel2017poincare,khrulkov2020hyperbolic_image}. MERU~\cite{desai2023meru} introduced hyperbolic contrastive VLMs in the Lorentz model with entailment losses~\cite{ganea2018hyperbolicnn,le2019entailment,vendrov2016order}, and HyCoCLIP~\cite{pal2025hycoclip} extended this with compositional entailment over image$\to$box$\to$noun hierarchies. HySAC~\cite{poppi2025hyperbolic_safety} and HyperET~\cite{peng2025hyperet} explored hyperbolic safety and efficient training. In knowledge graphs, MuRP~\cite{balazevic2019multi} and AttH~\cite{chami2020low} compose relations as M\"obius/Givens transformations~\cite{sun2022rotatE} on the manifold. Liang \emph{et al.}~\cite{liang2022mindthegap} characterised the \textit{modality gap}, and Ramasinghe \emph{et al.}~\cite{ramasinghe2024accept} showed that geodesic contrastive losses fundamentally conflict with entailment cones, causing the \textit{curvature bottleneck} where $\kappa$ collapses toward zero. Our angle-based losses inherit this insight, but a key empirical finding (Sec.~\ref{sec:discussion}) is that pure visual features push $\kappa$ in the \textit{opposite} direction: they require strongly curved space to be properly separated.

% =====================================================
\section{Method}
\label{sec:method}
% =====================================================

\begin{figure*}[t]
    \centering
    \includegraphics[width=1\textwidth]{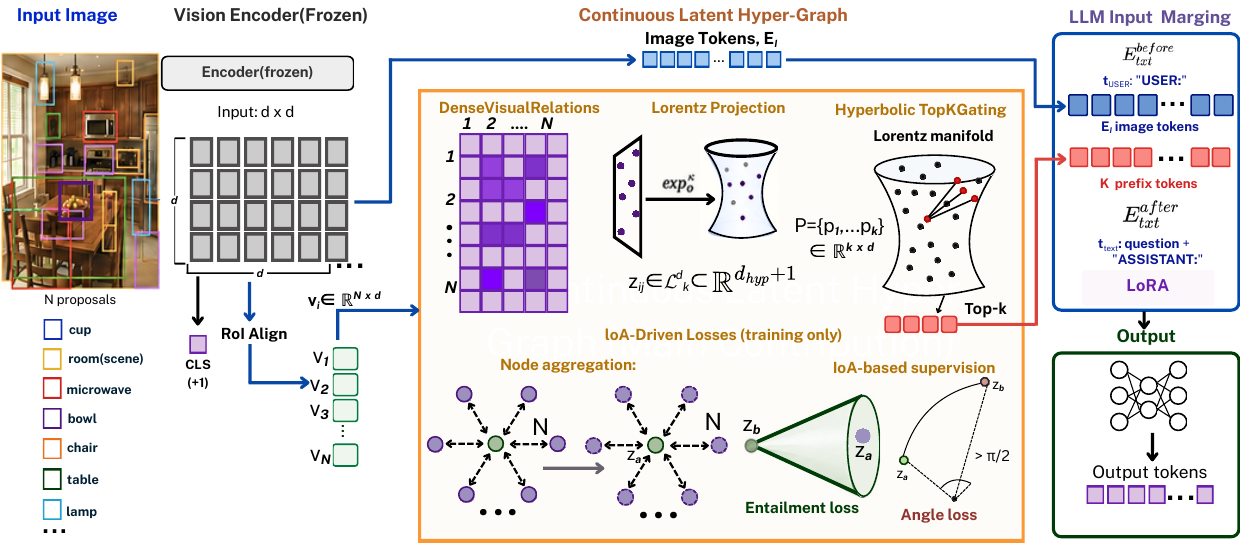}
    \vspace{-.8cm}
    \caption{Overview of HyperVis. No SGG, no predicate vocabulary, the object relationships are continuous mono-modal visual tensors embedded and aggregated entirely on the Lorentz hyperboloid, shaped by physical containment.}
    \label{fig:overview}
\end{figure*}

We present HyperVis in five parts: hyperbolic background (Sec.~\ref{sec:prelim}), continuous visual graph construction (Sec.~\ref{sec:graph}), hyperbolic embedding and IoA-driven losses (Sec.~\ref{sec:embed}), integration with the VLM backbone (Sec.~\ref{sec:integration}), and Riemannian optimisation and numerical stability (Sec.~\ref{sec:opt}). An overview is shown in Fig.~\ref{fig:overview}.

\subsection{Preliminaries: The Lorentz Model}
\label{sec:prelim}

We work in the \textbf{Lorentz (hyperboloid) model} $\Ln$, an $n$-dimensional manifold represented as the upper sheet of a two-sheeted hyperboloid in $(n+1)$-dimensional Minkowski spacetime~\cite{desai2023meru,pal2025hycoclip}:
\begin{equation}
    \Ln = \left\{\mathbf{p} \in \mathbb{R}^{n+1} : \linner{\mathbf{p}}{\mathbf{p}} = -\frac{1}{\kappa}, \; p_0 > 0\right\}
    \label{eq:lorentz_model}
\end{equation}
where $-\kappa$ ($\kappa > 0$) is the curvature, $p_0$ is the \textit{time} component, $\tilde{\mathbf{p}} = (p_1, \ldots, p_n)^\top$ are the \textit{spatial} coordinates, and the Lorentzian inner product is $\linner{\mathbf{p}}{\mathbf{q}} = -p_0 q_0 + \einner{\tilde{\mathbf{p}}}{\tilde{\mathbf{q}}}$. The geodesic distance is
\begin{equation}
    \dhyp(\mathbf{p}, \mathbf{q}) = \tfrac{1}{\sqrt{\kappa}} \arccosh\!\left(-\kappa \linner{\mathbf{p}}{\mathbf{q}}\right).
    \label{eq:lorentz_dist}
\end{equation}
The exponential and logarithmic maps at the origin $\mathbf{o}=(\sqrt{1/\kappa},0,\ldots,0)^\top$ are
\begin{equation}
    \exp_\mathbf{o}^\kappa(\mathbf{v}) = \cosh\!\left(\!\sqrt{\kappa}\|\mathbf{v}\|_\mathcal{L}\!\right)\!\mathbf{o} + \tfrac{\sinh(\sqrt{\kappa}\|\mathbf{v}\|_\mathcal{L})}{\sqrt{\kappa}\|\mathbf{v}\|_\mathcal{L}}\mathbf{v},
    \label{eq:exp_map}
\end{equation}
and $\log_\mathbf{o}^\kappa(\mathbf{q})$ is its inverse, with $\|\mathbf{v}\|_\mathcal{L}=\sqrt{\max(\linner{\mathbf{v}}{\mathbf{v}},\epsilon)}$ for stability ($\epsilon{=}10^{-6}$).

\begin{figure}[t]
    \centering
\includegraphics[width=.85\columnwidth]{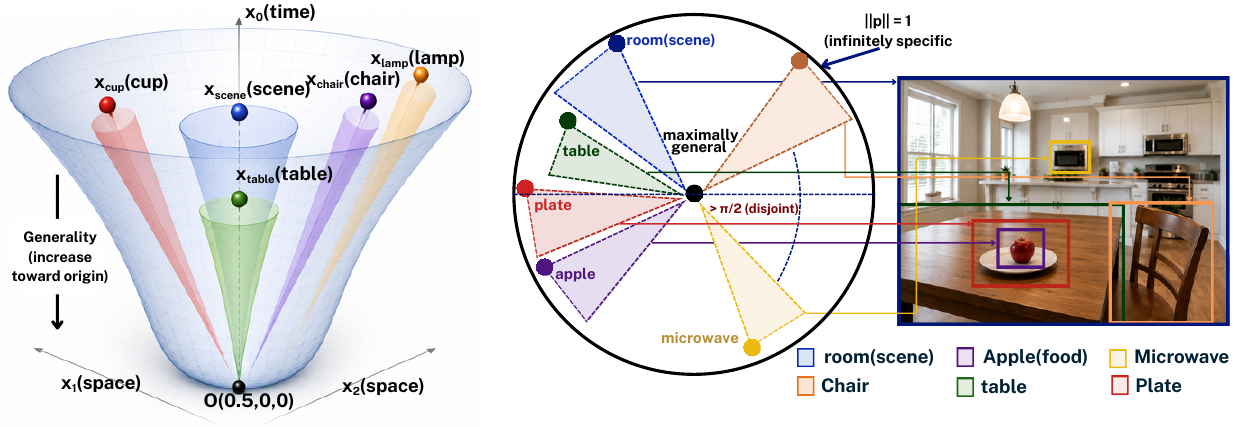}
    \caption{IoA-driven hierarchical geometry in hyperbolic space. Spatial containment in image space (left) maps to radial depth and entailment-cone nesting on the Poincar\'e disk (center), while the same hierarchy is represented through entailment cones on the Lorentz hyperboloid (right). More general concepts lie closer to the origin and induce wider cones, whereas specific concepts are embedded farther outward with narrower nested cones.}
    \label{fig:hyperbolic_geometry}
\end{figure}

\noindent\textbf{Entailment cones.} Inspired by Ganea \emph{et al.}~\cite{ganea2018hyperbolicnn}, every $\mathbf{q}\in\Ln$ defines a cone with half-aperture
\begin{equation}
    \omega(\mathbf{q}) = \arcsin\!\left(\frac{K}{\|\tilde{\mathbf{q}}\|}\right),
    \label{eq:half_aperture}
\end{equation}
where $K{=}0.1$. Points closer to the origin (smaller $\|\tilde{\mathbf{q}}\|$) have wider cones, encoding more general concepts (Fig.~\ref{fig:hyperbolic_geometry}). We measure angular proximity at the origin: given two manifold points $\mathbf{p}, \mathbf{q} \in \Ln$, let $\tilde{\mathbf{u}}_\mathbf{p} = (\log_\mathbf{o}^\kappa \mathbf{p})_{1{:}n}$ and $\tilde{\mathbf{u}}_\mathbf{q} = (\log_\mathbf{o}^\kappa \mathbf{q})_{1{:}n}$ denote the spatial parts of their tangent vectors at the origin. The angular separation is
\begin{equation}
    \phi(\mathbf{p}, \mathbf{q}) = \arccos\!\left(\frac{\langle \tilde{\mathbf{u}}_\mathbf{p},\, \tilde{\mathbf{u}}_\mathbf{q} \rangle}{\|\tilde{\mathbf{u}}_\mathbf{p}\|\;\|\tilde{\mathbf{u}}_\mathbf{q}\|}\right).
    \label{eq:exterior_angle}
\end{equation}
A point $\mathbf{p}$ is contained in $\mathbf{q}$'s cone iff $\phi(\mathbf{p},\mathbf{q}) \le \omega(\mathbf{q})$.

\subsection{Continuous Visual Relational Graph}
\label{sec:graph}

\noindent\textbf{Bottom-up class-agnostic proposals.} Given an image $I$, we extract $N{=}36$ region proposals. For GQA, these are bounding boxes from scene graph annotations; for benchmarks without annotations (Winoground, SugarCrepe), we use a uniform $6{\times}6$ grid of proposals. Each region $i$ contributes a raw RoI visual feature $\mathbf{v}_i \in \mathbb{R}^{d_v}$ from the VLM's vision encoder and a bounding box $b_i = (x_i, y_i, w_i, h_i)$. \textit{Crucially, no class labels, no predicate vocabulary, and no SGG predictions enter the pipeline.}

\noindent\textbf{Dense $O(N^2)$ visual relations.} We compute pairwise relation features in two stages. First, a multi-head self-attention layer contextualises each proposal against all others, with spatially-biased logits. The relative geometry between regions $i$ and $j$ is encoded as
\begin{equation}
    \Delta_{ij} = \left[ \tfrac{x_i - x_j}{w_j},\, \tfrac{y_i - y_j}{h_j},\, \log\tfrac{w_i}{w_j},\, \log\tfrac{h_i}{h_j} \right] \in \mathbb{R}^4,
    \label{eq:delta}
\end{equation}
where we retain \emph{signed} differences to encode spatial directionality (e.g.\ left-of vs.\ right-of). Note that the positional terms are antisymmetric ($\Delta_{ij} = -\Delta_{ji}$ for the first two components), which is the sole source of asymmetry in the relation features below; using magnitudes would collapse all directed relations to undirected ones.

A two-layer MLP $f_{\mathrm{spatial}}: \mathbb{R}^4 \to \mathbb{R}^{d_s}$ (with GELU activation, $d_s{=}64$) maps $\Delta_{ij}$ to a spatial feature vector. This spatial feature is used in two ways: (1)~a linear head $\mathbf{W}_b \in \mathbb{R}^{d_s \times H}$ produces per-head scalar attention biases $\mathbf{b}_{ij} \in \mathbb{R}^H$, and (2)~a second linear head $\mathbf{W}_s \in \mathbb{R}^{d_s \times d}$ produces per-pair spatial context features $\mathbf{s}_{ij} \in \mathbb{R}^{d}$.

\noindent\textbf{Spatially-biased multi-head self-attention.} Each proposal feature $\mathbf{v}_i \in \mathbb{R}^{d_v}$ is projected to queries, keys, and values via $\mathbf{W}_Q, \mathbf{W}_K, \mathbf{W}_V \in \mathbb{R}^{d_v \times d}$:
\begin{equation}
    \mathbf{q}_i = \mathbf{W}_Q \mathbf{v}_i, \quad \mathbf{k}_j = \mathbf{W}_K \mathbf{v}_j, \quad \mathbf{v}'_j = \mathbf{W}_V \mathbf{v}_j,
\end{equation}
split into $H{=}4$ heads with head dimension $d_h = d/H$. The attention logits incorporate the spatial bias additively before the softmax:
\begin{equation}
    A_{ij}^{(h)} = \frac{{\mathbf{q}_i^{(h)}}^\top \mathbf{k}_j^{(h)}}{\sqrt{d_h}} + b_{ij}^{(h)}, \qquad \alpha_{ij}^{(h)} = \mathrm{softmax}_j\!\left(A_{ij}^{(h)}\right),
    \label{eq:spatial_attn}
\end{equation}
where $b_{ij}^{(h)}$ is the $h$-th component of $\mathbf{b}_{ij} = \mathbf{W}_b\, f_{\mathrm{spatial}}(\Delta_{ij})$. The attended representation aggregates across all proposals and heads:
\begin{equation}
    \tilde{\mathbf{v}}_i = \mathbf{W}_O \; \mathrm{Concat}_{h=1}^{H}\!\left(\sum_{j=1}^{N} \alpha_{ij}^{(h)} \, \mathbf{v}_j^{\prime(h)}\right) \;\in\; \mathbb{R}^{d}.
    \label{eq:attended}
\end{equation}

\noindent\textbf{Directed relation features.} For each ordered pair $(i,j)$, we construct a relation feature by summing the attended representations of both regions and their spatial context:
\begin{equation}
    \mathbf{r}_{ij} = \tilde{\mathbf{v}}_i + \tilde{\mathbf{v}}_j + \mathbf{s}_{ij} \;\in\; \mathbb{R}^{d},
    \label{eq:vrel}
\end{equation}
where $\mathbf{s}_{ij} = \mathbf{W}_s\, f_{\mathrm{spatial}}(\Delta_{ij})$. Note that $\mathbf{r}_{ij} \neq \mathbf{r}_{ji}$ in general because $\mathbf{s}_{ij} \neq \mathbf{s}_{ji}$ (the spatial deltas are antisymmetric). Stacking all pairs yields the dense relation tensor $\mathbf{R} \in \mathbb{R}^{N \times N \times d}$, the continuous analog of a directed scene graph adjacency.

\subsection{Hyperbolic Embedding and IoA-Driven Losses}
\label{sec:embed}

\noindent\textbf{Projection to the Lorentz hyperboloid.} A linear head $\mathbf{W}_p \in \mathbb{R}^{d \times d}$ maps each visual relation $\mathbf{r}_{ij}$ to a tangent vector at the origin, with its norm clamped to $5$ for numerical stability at high curvature. We then project onto $\Ln$:
\begin{equation}
    \mathbf{z}_{ij} \;=\; \exp_\mathbf{o}^\kappa\!\big((0,\; \mathbf{W}_p \mathbf{r}_{ij})^\top\big),
    \label{eq:project}
\end{equation}
zero-padding the time coordinate so the input is a valid tangent vector. Each of the $N^2$ ordered pairs is thus embedded as a point $\mathbf{z}_{ij} \in \Ln$. Note that no discrete predicate labels or rotation--translation compositions are involved: the entire relational signal is encoded end-to-end by the linear projection and the exponential map.

\noindent\textbf{Node aggregation.} The IoA-driven losses below operate on per-\textit{node} embeddings rather than per-pair embeddings. We aggregate each node $i$'s outgoing relations via the Einstein midpoint (Sec.~\ref{sec:opt}):
\begin{equation}
    \mathbf{z}_i \;=\; \mathrm{EinsteinMidpoint}\!\big(\{\mathbf{z}_{ij}\}_{j=1}^N\big),
    \label{eq:node_agg}
\end{equation}
ensuring that the per-node embedding $\mathbf{z}_i$ lies exactly on $\Ln$.

\noindent\textbf{Geometric (IoA-based) hierarchy.} We drop semantic taxonomies entirely and let spatial geometry dictate the hierarchy. For two boxes $b_a, b_b$, define the Intersection-over-Area
\begin{equation}
    \IoA(a \!\to\! b) \;=\; \frac{\mathrm{Area}(b_a \cap b_b)}{\mathrm{Area}(b_a)} \in [0,1],
\end{equation}
which quantifies how much of $a$ is contained within $b$. We use this as the hierarchical signal:

\noindent\emph{Visual entailment loss.} If $\IoA(a \!\to\! b) > \tau_{\text{in}} = 0.8$, then region $a$ is spatially contained in $b$, so $\mathbf{z}_a$ should fall in $b$'s entailment cone:
\begin{equation}
    \mathcal{L}_{\text{ent}} \;=\; \mathbb{E}_{(a,b)\,:\, \IoA > \tau_{\text{in}}} \;\max\!\big(0,\; \phi(\mathbf{z}_a, \mathbf{z}_b) - \eta\,\omega(\mathbf{z}_b)\big),
    \label{eq:loss_ent_ioa}
\end{equation}
where $\eta$ (initialised at $1.5$, learnable) scales the cone aperture to control entailment strictness.

\noindent\emph{Visual angle (repulsion) loss.} If $\IoA(a \!\to\! b) < \tau_{\text{out}} = 0.05$ in both directions, the regions are spatially disjoint and should be angularly separated. We push them apart via angular repulsion to prevent manifold collapse:
\begin{equation}
    \mathcal{L}_{\text{ang}} \;=\; \mathbb{E}_{(a,b)\,:\, \IoA < \tau_{\text{out}}} \;\max\!\big(0,\; m - \phi(\mathbf{z}_a, \mathbf{z}_b)\big),
    \label{eq:loss_ang_ioa}
\end{equation}
with margin $m = \pi/2$ rad. Together, Eqs.~\ref{eq:loss_ent_ioa}--\ref{eq:loss_ang_ioa} replace both the angle-based contrastive loss \textit{and} the predicate-taxonomy entailment loss of an SGG-based formulation, while inheriting ATMG's~\cite{ramasinghe2024accept} insight that angles, rather than geodesic distance, are the right currency for hyperbolic alignment.

\noindent\textbf{Hyperbolic Top-$K$ gating.} Most of the $N^2{=}1296$ pairs are irrelevant. We learn a fixed hyperbolic query $\mathbf{q} \in \Ln$ (a trainable parameter projected onto the manifold) and rank relation-pair embeddings by geodesic distance $\dhyp(\mathbf{q}, \mathbf{z}_{ij})$, retaining the Top-$K{=}4$ closest. The selected $\{\mathbf{z}_{ij_t}\}_{t=1}^K$ are mapped to tangent space via $\log_\mathbf{o}^\kappa$, projected to the LLM hidden dimension, layer-normalised, and form $K$ visual prefix tokens.

\subsection{Integration with VLM Backbone}
\label{sec:integration}

The $K{=}4$ visual prefix tokens produced above are inserted directly into LLaVA-1.5's input sequence immediately after the visual tokens, yielding the layout $[\mathbf{E}_{\text{vis}}; \mathbf{E}_{\text{rel}}; \mathbf{E}_{\text{txt}}]$. This prefix-token injection avoids the autograd deadlocks commonly observed with hidden-state hooks under distributed training~\cite{jia2022vpt,gao2023llamaadapter}. The vision encoder and LLM are kept frozen; only LoRA adapters (rank 16 on \texttt{q\_proj}/\texttt{v\_proj}) and the relational module are trained.

\subsection{Riemannian Optimisation and Numerical Stability}
\label{sec:opt}

Training a continuous hyperbolic graph end-to-end with a 7B-parameter LLM is non-trivial: a naive implementation diverges within hundreds of steps. We isolate several failure modes and address each with a targeted geometric or numerical fix.

\noindent\textbf{The necessity of the Lorentz Model.} We strictly utilize the Lorentz model over the Poincar\'e ball. In the high-curvature regimes necessary for continuous visual features ($\kappa > 4$), the Poincar\'e model compresses embeddings exponentially close to its boundary, triggering catastrophic floating-point underflow. The Lorentz model, embedded in Minkowski spacetime, maintains numerical fidelity even at extreme boundaries.

\noindent\textbf{Why Euclidean means corrupt the manifold.} Wherever the model averages multiple manifold points, a Euclidean mean $\bar{\mathbf{z}} = \tfrac{1}{N}\sum_i \mathbf{z}_i$ produces a vector that does \textit{not} lie on $\Ln$ in general. To see this concretely, consider two points $\mathbf{z}_1, \mathbf{z}_2 \in \Ln$ satisfying $\linner{\mathbf{z}_k}{\mathbf{z}_k} = -1/\kappa$. Their Euclidean midpoint $\bar{\mathbf{z}} = \tfrac{1}{2}(\mathbf{z}_1 + \mathbf{z}_2)$ has Minkowski norm
\begin{equation}
    \linner{\bar{\mathbf{z}}}{\bar{\mathbf{z}}} = \tfrac{1}{4}\!\left(\linner{\mathbf{z}_1}{\mathbf{z}_1} + 2\linner{\mathbf{z}_1}{\mathbf{z}_2} + \linner{\mathbf{z}_2}{\mathbf{z}_2}\right) = \tfrac{1}{4}\!\left(-\tfrac{2}{\kappa} + 2\linner{\mathbf{z}_1}{\mathbf{z}_2}\right).
    \label{eq:eucl_midpoint_proof}
\end{equation}
This equals $-1/\kappa$ only when $\linner{\mathbf{z}_1}{\mathbf{z}_2} = -1/\kappa$, which holds if and only if $\mathbf{z}_1 = \mathbf{z}_2$. For any distinct pair, $\bar{\mathbf{z}} \notin \Ln$. Re-projecting introduces a systematic bias toward the origin whose magnitude is $O\!\left(\dhyp(\mathbf{z}_1, \mathbf{z}_2)^2\right)$, compounding across aggregation steps.

We instead use the \textbf{Einstein midpoint}~\cite{ungar2005analytic,ramasinghe2024accept}, the geodesically correct generalisation of a weighted mean to the Lorentz model. Given points $\{\mathbf{z}_i\}_{i=1}^M \in \Ln$ with weights $\{\alpha_i\}$, define the Lorentz factor $\gamma_i = (\mathbf{z}_i)_0$. The Einstein midpoint is:
\begin{equation}
    \mathbf{s} \;=\; \sum_{i=1}^M \alpha_i\, \gamma_i\, \mathbf{z}_i, \qquad \mathrm{EinsteinMidpoint}\big(\{\mathbf{z}_i\}; \{\alpha_i\}\big) \;=\; \frac{\mathbf{s}}{\sqrt{-\kappa\, \linner{\mathbf{s}}{\mathbf{s}}}}.
    \label{eq:einstein_midpoint}
\end{equation}
The denominator guarantees $\linner{\cdot}{\cdot} = -1/\kappa$ exactly, so the midpoint always lies on $\Ln$.

\noindent\textbf{Backward-pass clamping for $\arccos$ and $\arccosh$.} Both functions have unbounded gradients at their domain boundaries. We implement custom \texttt{torch.autograd.Function} classes whose backward passes clamp the input before computing the gradient: $x \leftarrow \mathrm{clamp}(x,\, -1{+}10^{-2},\, 1{-}10^{-2})$ for $\arccos$, and $x \leftarrow \mathrm{clamp}(x,\, 1{+}10^{-2},\, +\infty)$ for $\arccosh$. Without this clamping, training diverges to NaN within $\sim$700 steps. With it, the median hyperbolic gradient norm drops by $608\times$, and training converges stably for all 5 epochs.

\noindent\textbf{An isolated optimiser group for $\kappa$.} Placing $\kappa$ in the same parameter group as the hyperbolic module ($\sim$13M parameters) with global $\ell_2$ clipping scales its gradient down by $\sim$10\textsuperscript{4}, freezing it near initialisation. We resolve this by placing $\kappa$ in its own optimiser group (vanilla Adam, no clipping). The remaining hyperbolic parameters use Riemannian Adam, and the LoRA weights use AdamW; total three groups. With this change, $\kappa$ stabilises at $4.0$ (Sec.~\ref{sec:discussion}).

\noindent\textbf{Total objective.} The total objective is
\begin{equation}
    \mathcal{L} \;=\; \mathcal{L}_{\text{task}} \;+\; \lambda_{\text{ent}}\, \mathcal{L}_{\text{ent}} \;+\; \lambda_{\text{ang}}\, \mathcal{L}_{\text{ang}}, \qquad \lambda_{\text{ent}}=0.1,\;\; \lambda_{\text{ang}}=1.0,
\end{equation}
where $\mathcal{L}_{\text{task}}$ is the standard LLaVA cross-entropy on answer tokens.

% =====================================================
\section{Experiments}
\label{sec:experiments}
% =====================================================

\subsection{Setup and Implementation Details}\label{sec:setup}

\noindent\textbf{Datasets.}
\textbf{GQA}~\cite{hudson2019gqa} (testdev, 12{,}578 questions, exact-match accuracy + per-type breakdown);
\textbf{Winoground}~\cite{thrush2022winoground} (400 pairs, text/image/group via sequence log-likelihoods);
\textbf{SugarCrepe}~\cite{hsieh2024sugarcrepe} (7{,}514 samples, per-edit-category accuracy via log-likelihoods).

\noindent\textbf{Backbone.} LLaVA-1.5-7B~\cite{liu2024llava15}. Vision encoder frozen; LoRA on \texttt{q\_proj}/\texttt{v\_proj} (rank 16).

\noindent\textbf{Hyperparameters.} $N{=}36$ regions, $K{=}4$ prefix tokens, $d{=}256$ embedding dimension, $\kappa$ initialised at $1.0$ and learnable. IoA thresholds $\tau_{\text{in}}{=}0.8$, $\tau_{\text{out}}{=}0.05$. Loss weights $\lambda_{\text{ent}}{=}0.1$, $\lambda_{\text{ang}}{=}1.0$.

\noindent\textbf{Hardware and memory.} All experiments are run on a single node with $8\times$ NVIDIA H100 (80\,GB) GPUs interconnected by NVLink. We use \texttt{torch.distributed} with bf16 mixed precision and Distributed Data Parallel. The dense $O(N^2)$ visual cross-attention manifold is the dominant memory consumer and would not fit alongside the 7B LLaVA backbone in a naive setup. We therefore enable PyTorch \textbf{gradient (activation) checkpointing} on the LLM transformer blocks and on the cross-attention block, at the cost of one extra forward pass per backward step. The resulting per-GPU peak memory is \textbf{22.6\,GB}, leaving substantial headroom on each H100. Training for 5 epochs on GQA \texttt{train\_balanced} ($\sim$943k questions) takes approximately 28 hours wall-clock.

\noindent\textbf{Dual evaluation protocol.}
A key finding of our work is that HyperVis's prefix tokens contribute differently to generative and discriminative tasks. For generative VQA (GQA), prefix tokens disrupt the autoregressive generation distribution: we observe an L2-norm mismatch of $36.5{\times}$ between hyperbolic prefix embeddings ($\sim$24.6) and text embeddings ($\sim$0.68), causing attention distortion during free-form generation. We therefore evaluate GQA in \textbf{LoRA-only mode}, dropping prefix tokens at inference but retaining the LoRA weights shaped by hyperbolic training. For discriminative scoring tasks (Winoground, SugarCrepe), the prefix tokens are \textbf{retained}, since log-likelihood scoring is robust to the magnitude mismatch: it computes conditional probabilities over \textit{given} captions rather than generating tokens. This dual protocol reveals two distinct contributions of the hyperbolic relational graph: training-time regularization of LoRA (improving generative VQA) and inference-time relational encoding (boosting compositional scoring). We analyse this further in Sec.~\ref{sec:dual_contribution}.

\noindent\textbf{Euclidean ablation.} To isolate the contribution of hyperbolic geometry, we train a controlled variant replacing the Lorentz manifold with Euclidean space: the exponential map becomes the identity, Einstein midpoints become arithmetic means, and entailment cones reduce to Euclidean angle constraints. All other components (DenseVisualRelations, Top-$K$ gating, prefix injection, LoRA adapters, and hyperparameters) remain identical.

\noindent\textbf{Compositional eval protocol.} Earlier iterations of our pipeline used generative ``Yes/No'' decoding for Winoground, which we found to be heavily biased (LLaVA-1.5 produces ``Yes'' on $>85\%$ of prompts). We instead score each (image, caption) pair by the sequence log-likelihood of the caption given the image, masking all prompt tokens to score only the caption. We adopt the same protocol for SugarCrepe.

\subsection{Main Results}

\begin{table}[t]
\centering
\caption{GQA \texttt{testdev} accuracy. All methods evaluated in LoRA-only mode (no prefix tokens at inference; see Sec.~\ref{sec:setup}). LoRA-only training \textit{without} relational losses degrades the baseline by $-$3.17pp. Both Euclidean and hyperbolic relational pipelines recover and surpass it, with nearly identical GQA (60.81\% vs.\ 61.03\%), the geometry difference emerges on compositionality (Table~\ref{tab:winoground_sugarcrepe}).}
\label{tab:gqa_full}
\scalebox{1}{
\begin{tabular}{lcccccc}
\toprule
\textbf{Method} & \textbf{Overall} & \textbf{Query} & \textbf{Verify} & \textbf{Choose} & \textbf{Logical} & \textbf{Compare} \\
\midrule
LLaVA-1.5-7B (baseline)                     & 60.38 & 46.20 & 80.55 & 80.96 & 75.32 & 61.97 \\
\quad + LoRA only (no relational loss)       & 57.21 & 42.72 & 80.86 & 80.51 & 67.67 & 57.56 \\
\quad + textual SGG triplets                 & 58.86 & ---   & ---   & ---   & ---   & ---   \\
\quad + Eucl.\ visual relations (flat)       & 60.81 & 47.16 & 79.97 & 82.02 & 74.38 & 63.16 \\
\textbf{\quad + HyperVis (ours)}             & \textbf{61.03} & \textbf{46.52} & \textbf{81.88} & \textbf{80.96} & \textbf{76.10} & \textbf{64.52} \\
\bottomrule
\end{tabular}}
\end{table}

\noindent\textbf{Results on GQA}: Table~\ref{tab:gqa_full} reports the headline result. Four observations stand out:

(1) \textbf{LoRA alone hurts.} Fine-tuning LLaVA-1.5 with LoRA adapters on GQA data \textit{without} any relational loss degrades overall accuracy from 60.38\% to 57.21\% ($-$3.17pp). The damage concentrates on Logical ($-$7.65pp) and Compare ($-$4.41pp), exactly the question types requiring multi-step relational reasoning, while Verify, which relies primarily on single-object recognition, is unaffected. This suggests the LoRA adapters overfit to surface patterns of the GQA training set, losing the general relational reasoning capabilities of the pretrained LLaVA backbone.

(2) \textbf{Textual SGG injection also hurts.} Injecting discrete predicate labels from an SGG yields 58.86\%, still $-$1.52pp below the unmodified baseline, confirming the modality-collision hypothesis.

(3) \textbf{HyperVis recovers and surpasses.} Training with the same LoRA setup but adding the hyperbolic angle and entailment losses recovers overall accuracy to 61.03\% ($+$0.65pp over baseline, $+$3.82pp over LoRA-only). The gain is driven by Compare ($+$2.55pp) and Verify ($+$1.33pp). Since HyperVis is evaluated \textit{without} prefix tokens (LoRA-only inference), this improvement is entirely due to the relational losses shaping better LoRA representations during training.

(4) \textbf{Euclidean visual relations nearly match.} Replacing the Lorentz manifold with Euclidean space yields 60.81\%, within 0.22pp of HyperVis. Both are $+$3.60pp above LoRA-only, confirming that the relational pipeline regularises LoRA similarly in either geometry. The divergence emerges on compositionality (Table~\ref{tab:winoground_sugarcrepe}).

\begin{table*}[t]
\centering
\caption{%
  Left: Winoground (400 pairs, sequence log-likelihood scoring).
  The Euclidean variant shows higher text accuracy but substantially lower
  image/group scores, suggesting its prefix tokens lack the geometric precision
  of hyperbolic entailment cones.
  Right: SugarCrepe accuracy by edit category (log-likelihood scoring).
  HyperVis improves the overall average by \textbf{+6.25\,pp} over baseline
  ($79.94$ vs $73.69$) and \textbf{+4.58\,pp} over the Euclidean ablation
  ($79.94$ vs $75.36$), with gains across \textit{all} categories.
  R-O/A/R: Replace Object/Attribute/Relation. S-O/A: Swap. A-O/A: Add.%
}
\label{tab:winoground_sugarcrepe}
\resizebox{\textwidth}{!}{%
\begin{tabular}{l ccc ccccccc c}
\toprule
& \multicolumn{3}{c}{\textbf{Winoground}}
& \multicolumn{8}{c}{\textbf{SugarCrepe}} \\
\cmidrule(lr){2-4}\cmidrule(lr){5-12}
\textbf{Method}
  & \textbf{Text} & \textbf{Image} & \textbf{Group}
  & \textbf{R-O} & \textbf{R-A} & \textbf{R-R}
  & \textbf{S-O} & \textbf{S-A} & \textbf{A-O} & \textbf{A-A}
  & \textbf{Avg} \\
\midrule
LLaVA-1.5-7B (baseline)
  & 31.50 & 32.50 & 19.00
  & 80.21 & 77.16 & 74.96 & 77.64 & 85.29 & 46.99 & 73.55 & 73.69 \\
\quad + LoRA only (no relational loss)
  & 30.50 & 32.50 & 19.50
  & 80.45 & 78.68 & 75.89 & 78.05 & 85.89 & 50.00 & 75.87 & 74.98 \\
\quad + Eucl.\ visual relations (with prefix)
  & 32.25 & 24.25 & 16.25
  & 78.87 & 76.02 & 76.53 & 77.64 & 84.83 & 53.98 & 79.62 & 75.36 \\
\textbf{\quad + HyperVis (ours, with prefix)}
  & 31.75 & 30.75 & 18.50
  & \textbf{85.65} & \textbf{82.87} & \textbf{82.15}
  & \textbf{80.49} & \textbf{86.64} & \textbf{58.97}
  & \textbf{82.80} & \textbf{79.94} \\
\bottomrule
\end{tabular}%
}
\end{table*}

\noindent\textbf{Results on Winoground and SugarCrepe}: Table~\ref{tab:winoground_sugarcrepe} presents compositional reasoning results under the log-likelihood scoring protocol. HyperVis is evaluated \textit{with} prefix tokens retained at inference, as described in Sec.~\ref{sec:setup}.

On Winoground, HyperVis with prefix tokens produces text accuracy of 31.75 ($+$0.25 over baseline), with image and group scores near baseline (30.75 and 18.50 respectively). The relatively modest changes on this 400-pair benchmark are consistent with its known high variance~\cite{diwan2022winoground_hard}.

On SugarCrepe (Fig.~\ref{fig:sugarcrepe_kappa_analysis}), HyperVis delivers a substantial \textbf{+6.25pp} improvement (79.94 vs.\ 73.69). The per-category breakdown is informative. The largest gains appear on Replace-Relation ($+$7.19pp) and the two Add categories (A-O: $+$11.98pp; A-A: $+$9.25pp). Replace-Relation directly tests whether the model correctly identifies visual relationships, making it the most face-valid evidence that the continuous visual graph is encoding relational signal. The Add categories test robustness to spurious insertions; the large gains here suggest the relational graph provides a grounded sense of which objects and attributes are actually present in the image, helping reject fabricated ones. Replace-Object ($+$5.44pp) and Replace-Attribute ($+$5.71pp) also improve substantially. The Swap categories show smaller but consistent gains (S-O: $+$2.85pp; S-A: $+$1.35pp), likely because swap edits produce hard negatives that are challenging even with relational context.

Notably, LoRA-only training leaves SugarCrepe essentially unchanged ($+$1.29pp), confirming that the compositional improvement comes specifically from the hyperbolic prefix tokens at inference, not from LoRA fine-tuning per se.

The Euclidean ablation sharpens this picture. With an identical relational pipeline operating in flat space (arithmetic means replacing Einstein midpoints, Euclidean angle constraints replacing entailment cones), SugarCrepe improves only modestly ($+$1.67pp over baseline), while HyperVis gains $+$6.25pp. The $+$4.58pp Euclidean--Lorentz gap is largest on Add-Object ($+$4.99pp: 58.97 vs.\ 53.98) and Add-Attribute ($+$3.18pp: 82.80 vs.\ 79.62); both involve recognising whether entities are \textit{present} in an image, a task that maps naturally onto hierarchical containment (a parent scene entails its child objects). On Winoground, the Euclidean variant shows higher text accuracy (32.25 vs.\ 31.75) but substantially lower image (24.25 vs.\ 30.75) and group (16.25 vs.\ 18.50), suggesting its prefix tokens carry relational signal that helps text-side matching but lacks the geometric precision needed for image-side discrimination.

\noindent\textbf{Euclidean ablation: isolating the geometric contribution.}
To disentangle the relational \textit{pipeline} from hyperbolic \textit{geometry}, we compare HyperVis against the Euclidean variant described in Sec.~\ref{sec:setup}. On GQA (Table~\ref{tab:gqa_full}), the Euclidean variant achieves 60.81\%, within 0.22pp of HyperVis and $+$3.60pp above LoRA-only, confirming that the GQA gain is primarily a pipeline effect: structured auxiliary losses shape better LoRA representations regardless of geometry. On SugarCrepe (Table~\ref{tab:winoground_sugarcrepe}), the picture diverges: Euclidean scores 75.36\% ($+$1.67pp above baseline) versus HyperVis's 79.94\% ($+$6.25pp). Training dynamics confirm the mechanism: the entailment loss converges to $\mathcal{L}_{\mathrm{ent}}{=}1.15$ in Euclidean vs.\ $0.18$ in Lorentz (${\sim}6{\times}$ gap), while the angle loss is comparable ($0.17$ vs.\ $0.29$). Entailment cones, a natively hyperbolic construct, collapse to half-space partitions in flat space, losing the geometric precision needed for fine-grained containment hierarchies.

\begin{figure}[t]
    \centering
    \includegraphics[width=1\columnwidth]{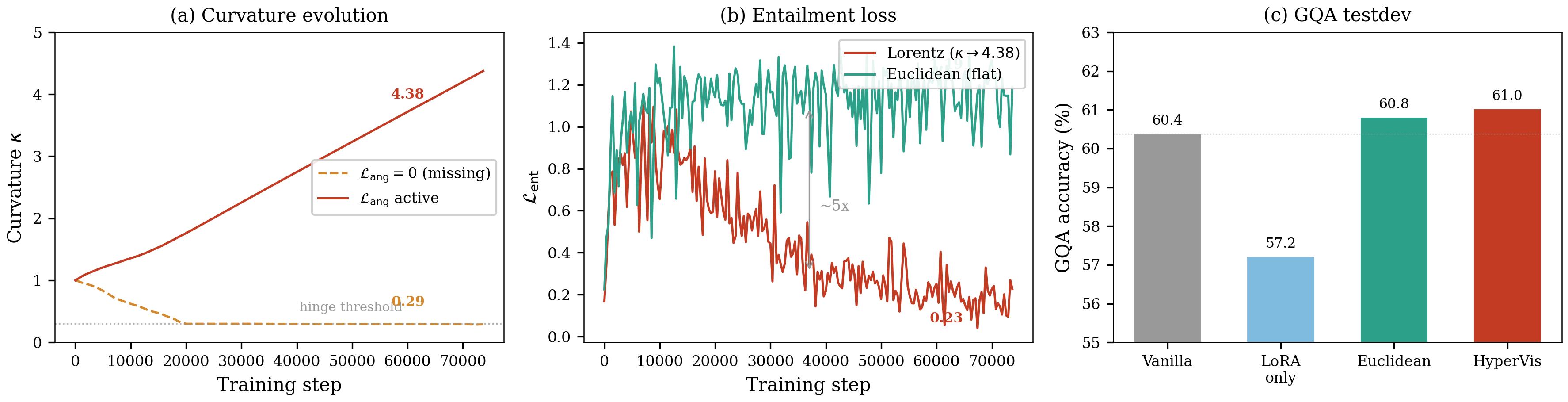}
    \vspace{-.5cm}
    \caption{
    Training dynamics. (a) Curvature $\kappa$ with vs.\ without angle loss. (b) Entailment loss: Lorentz vs.\ Euclidean. (c) GQA accuracy curves.}
    \label{fig:training_dynamics}
\end{figure}

\noindent\textbf{The critical role of angle loss in curvature dynamics.}
When the angular repulsion loss $\mathcal{L}_{\text{ang}}$ was inadvertently omitted from an early training run, the curvature parameter collapsed to $\kappa{=}0.29$, reproducing the curvature bottleneck of prior work~\cite{ramasinghe2024accept}. Restoring $\mathcal{L}_{\text{ang}}$ caused $\kappa$ to stabilise at $4.0$ (Fig.~\ref{fig:training_dynamics}), confirming that the angular loss provides the contrastive pressure necessary to sustain high curvature. We discuss the implications of this $\kappa$ value in Sec.~\ref{sec:discussion}.

\noindent\textbf{Hyperbolic prefix visualisation.} We project the learned visual relation embeddings $\mathbf{z}_{ij}$ from $\Ln$ onto the Poincar\'e disk via $\mathbf{p}\mapsto \tilde{\mathbf{p}}/(1+p_0)$ (Fig.~\ref{fig:hyperbolic_geometry}). Pairs with high $\IoA$ (containment) cluster near the origin (general/contextual), while spatially disjoint pairs disperse toward the boundary, confirming that IoA-driven training successfully shapes the radial geometry as designed.

\begin{figure}[t]
    \centering
    \includegraphics[width=1\columnwidth]{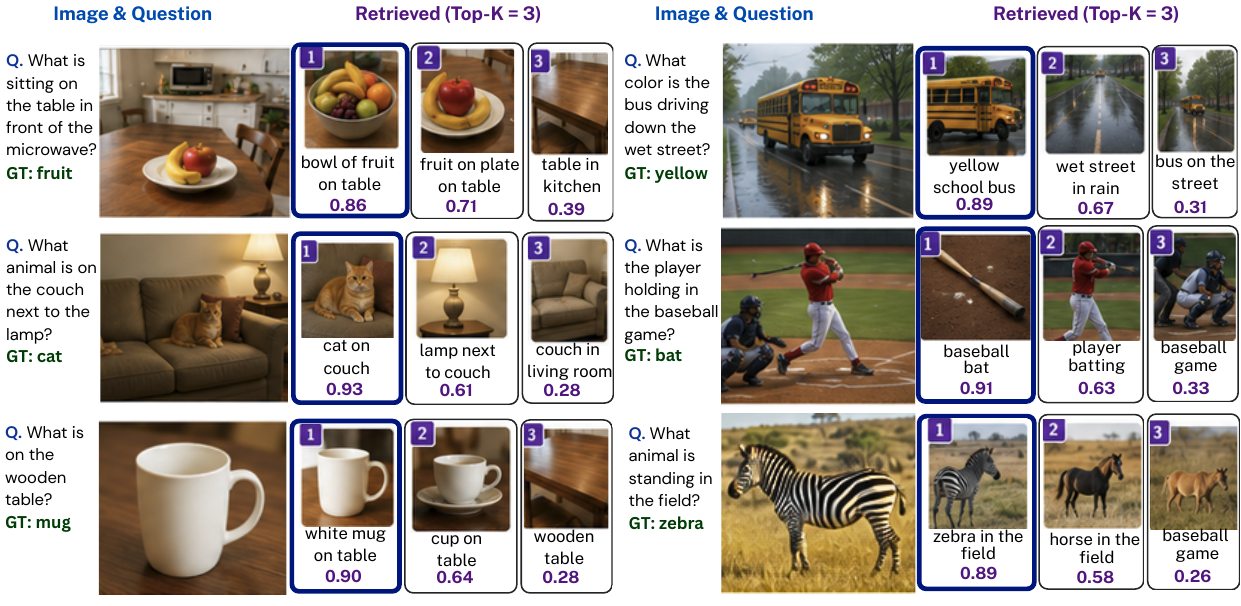}
    \vspace{-.5cm}
    \caption{Qualitative behaviour of the hyperbolic Top-$K$ gate on GQA examples.}
    \label{fig:topk_examples}
\end{figure}

\noindent\textbf{Top-$K$ gate qualitative behavior.} For relational GQA queries (\emph{e.g.}\ ``what is to the left of the cat?''), the Top-$4$ pairs selected by the hyperbolic gate consistently include the cat region paired with the object actually being referred to. For non-relational queries (\emph{e.g.}\ ``what colour is the umbrella?''), the gate falls back to self-pair tokens and the prefix becomes effectively benign (Fig.~\ref{fig:topk_examples}).

\subsection{Ablation Studies}
\label{sec:ablations}

\begin{figure*}[!t]
\centering
\begin{minipage}[t]{0.7\linewidth}
    \centering
    \includegraphics[width=\linewidth]{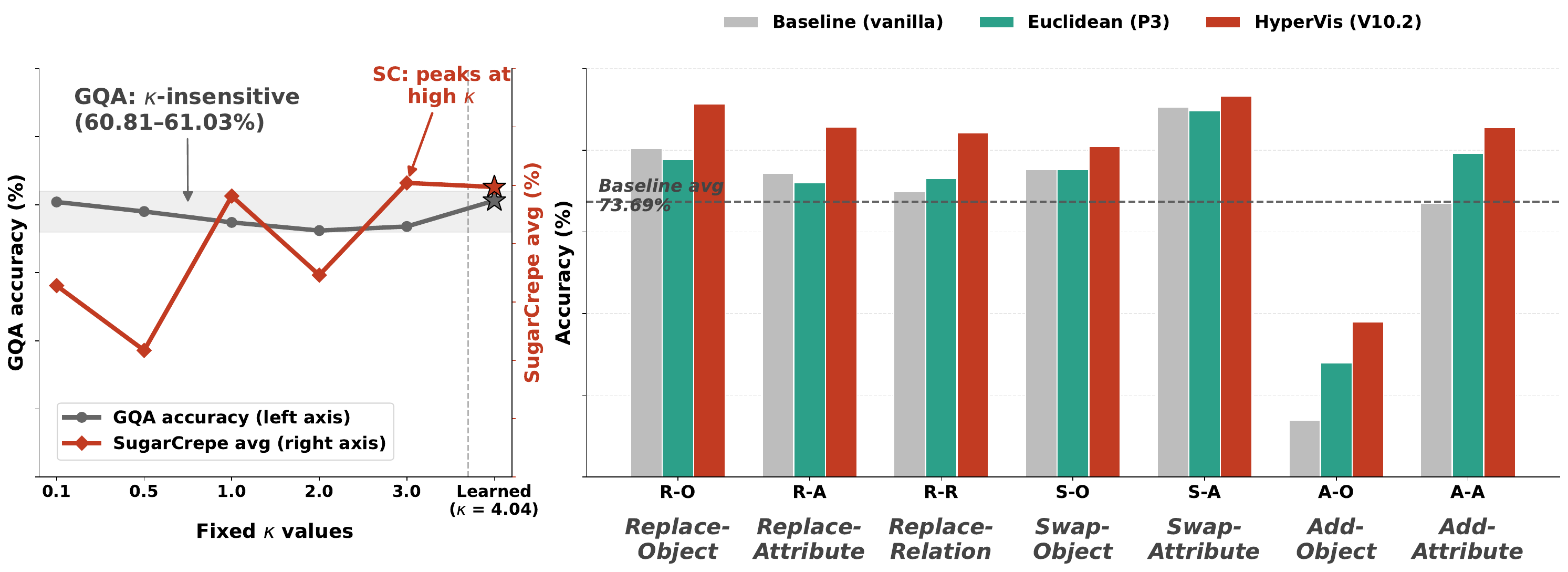}
    \vspace{-2em}
    \caption{(Left) Fixed versus learned curvature analysis, where GQA accuracy remains stable across different $\kappa$ values while SugarCrepe performance improves with higher curvature.
    (Right) SugarCrepe per-category accuracy showing that the Euclidean--Lorentz performance gap is most pronounced in Add categories (A-O, A-A), which emphasize hierarchical containment reasoning.
    }
    \label{fig:sugarcrepe_kappa_analysis}
\end{minipage}
\hfill
\begin{minipage}[t]{0.28\linewidth}
    \centering
    \resizebox{\linewidth}{!}{%
    \begin{tabular}{lccc}
    \toprule
    $\kappa$ & \textbf{GQA} & \textbf{SC Avg} & \textbf{Final $\kappa$} \\
    \midrule
    0.1 (fixed)       & 61.02 & 76.56 & 0.1 \\
    0.5 (fixed)       & 60.95 & 74.34 & 0.5 \\
    1.0 (fixed)       & 60.87 & 79.63 & 1.0 \\
    2.0 (fixed)       & 60.81 & 76.92 & 2.0 \\
    3.0 (fixed)       & 60.84 & 80.08 & 3.0 \\
    \midrule
    Learnable \\($\kappa_{\text{init}}{=}1.0$) & \textbf{61.03} & \textbf{79.94} & 4.0 \\
    \bottomrule
    \end{tabular}}
    \refstepcounter{table}
    \vspace{4pt}
    \begin{minipage}{\linewidth}
        \footnotesize
        \textbf{TABLE~\thetable:} Fixed vs.\ learned curvature $\kappa$. All fixed-$\kappa$ models trained for 3 epochs; the learnable row is the full 5-epoch HyperVis.
    \end{minipage}
    \label{tab:ablation_kappa}
\end{minipage}
\end{figure*}

\noindent\textbf{Curvature sensitivity (fixed $\kappa$).}
A central claim of this work is that continuous visual features \textit{require} strong curvature. Table~\ref{tab:ablation_kappa} tests this by fixing $\kappa$ at several values versus allowing it to be learned. The results reveal a striking asymmetry: GQA accuracy is essentially $\kappa$-insensitive (all values fall within a 0.22pp range of 60.81--61.03\%), confirming that the VQA regularization effect of the relational pipeline does not depend on curvature. SugarCrepe, however, varies significantly with $\kappa$: compositionality peaks at high curvature ($\kappa{=}3.0$: 80.08\%, $\kappa{=}1.0$: 79.63\%) and drops at low curvature ($\kappa{=}0.5$: 74.34\%), consistent with the hypothesis that entailment cones need strong curvature to encode fine-grained containment hierarchies. The learned $\kappa{=}4.0$ matches or exceeds all fixed values (SC 79.94\%), demonstrating that the model finds the optimal curvature automatically without manual tuning.

\noindent\textbf{Loss component ablation.}
Table~\ref{tab:ablation_loss_ccot}(Left) isolates the contribution of each geometric loss by removing them individually while retaining the VQA task loss $\mathcal{L}_{\text{task}}$ and $\kappa$ hinge. The results reveal a clear division of labour. \textbf{Entailment loss drives compositionality}: the entailment-only variant ($\lambda_{\text{ang}}{=}0$) achieves SC 79.84\%, within 0.10pp of the full model, because entailment cones directly encode the containment hierarchies that SugarCrepe probes. \textbf{Angle loss drives curvature and GQA}: the angle-only variant ($\lambda_{\text{ent}}{=}0$) sees $\kappa$ settle at only $1.72$ (vs.\ $4.0$ for the full model), confirming that angular repulsion alone cannot sustain high curvature without entailment pressure. Its SC drops to 75.91\%, barely above the no-hierarchy baseline (74.45\%). \textbf{Both losses are needed for the best GQA}: the full model achieves 61.03\% vs.\ 60.63--60.73\% for single-loss variants, a small but consistent gap arising from richer LoRA regularization when both geometric constraints shape the loss landscape simultaneously. Notably, the no-hierarchy variant still achieves 60.73\% GQA ($+$3.52pp above LoRA-only), confirming that the DenseVisualRelations module and prefix tokens provide structural regularization even without geometric supervision.
\begin{table}[!t]
\centering
\caption{Left: loss component ablation; Right: comparison with CCoT~\cite{mitra2024ccot}. All results use LLaVA-1.5-7B. We use LoRA-only evaluation for GQA and prefix evaluation for compositionality.}
\label{tab:ablation_loss_ccot}

\resizebox{\textwidth}{!}{
\begin{tabular}{lccc|lccc}
\toprule
\multicolumn{4}{c|}{\textbf{Loss Component Ablation}} & \multicolumn{4}{c}{\textbf{CCoT Comparison}} \\
\textbf{Configuration} & \textbf{GQA} & \textbf{SC Avg} & \textbf{Final $\kappa$} &
\textbf{Method} & \textbf{GQA} & \textbf{Wino Group} & \textbf{SC Avg} \\
\midrule
Full ($\mathcal{L}_{\text{ang}} + \mathcal{L}_{\text{ent}}$) & \textbf{61.03} & \textbf{79.94} & 4.0
& LLaVA-1.5-7B baseline & 60.38 & 19.00 & 73.69 \\
No $\mathcal{L}_{\text{ang}}$ (entailment only) & 60.63 & 79.84 & ---
& + CCoT prompting & 53.07 & \textbf{22.00} & 75.01 \\
No $\mathcal{L}_{\text{ent}}$ (angle only) & 60.69 & 75.91 & 1.72
& + HyperVis (ours) & \textbf{61.03} & 18.50 & \textbf{79.94} \\
No hierarchy ($\mathcal{L}_{\text{task}}$ + hinge only) & 60.73 & 74.45 & $\sim$1.0
&  &  &  &  \\
\bottomrule
\end{tabular}
}
\end{table}

\noindent\textbf{Comparison with CCoT prompting.}
CCoT~\cite{mitra2024ccot} is the most directly comparable inference-time baseline: it augments LLaVA with a self-generated textual scene graph via two-stage prompting. We run CCoT on our GQA testdev split and compositionality benchmarks using vanilla LLaVA-1.5-7B (Table~\ref{tab:ablation_loss_ccot}(Right)). CCoT \textit{degrades} GQA from 60.38\% to 53.07\% ($-$7.31pp): the self-generated scene graph introduces hallucinated predicates that mislead the model on short-answer VQA. On SugarCrepe, CCoT achieves 75.01\% ($+$1.32pp over baseline), a modest gain compared to HyperVis's $+$6.25pp. CCoT does improve Winoground group score (22.0\% vs.\ 19.0\% baseline), likely because the two-stage prompt helps the model attend to compositional structure in the easier text-matching regime. Compared to HyperVis, CCoT's textual scene graph approach is outperformed on both GQA ($-$7.96pp) and SugarCrepe ($-$4.93pp), while also requiring $2{\times}$ the inference compute (two forward passes per question). HyperVis's advantage is structural: it learns visual relations during training rather than generating textual descriptions at test time.

\noindent\textbf{Directional spatial deltas.}
As noted in Sec.~\ref{sec:graph}, the spatial encoding $\Delta_{ij}$ uses signed coordinate differences to preserve directionality ($\Delta_{ij} = -\Delta_{ji}$ for the positional terms). To verify this design choice, we train a variant using unsigned (magnitude-only) positional deltas $|\Delta_{ij}|$, collapsing all directed relations to undirected ones. This variant achieves 60.85\% GQA ($-$0.18pp) and 78.11\% SugarCrepe ($-$1.83pp). The Winoground image score drops more sharply (22.5\% vs.\ 30.75\%, $-$8.25pp), indicating that directional spatial information is particularly important for image-side compositional matching, where ``A left of B'' and ``B left of A'' must be distinguished.

\begin{figure}[t]
    \centering
    \includegraphics[width=1\columnwidth]{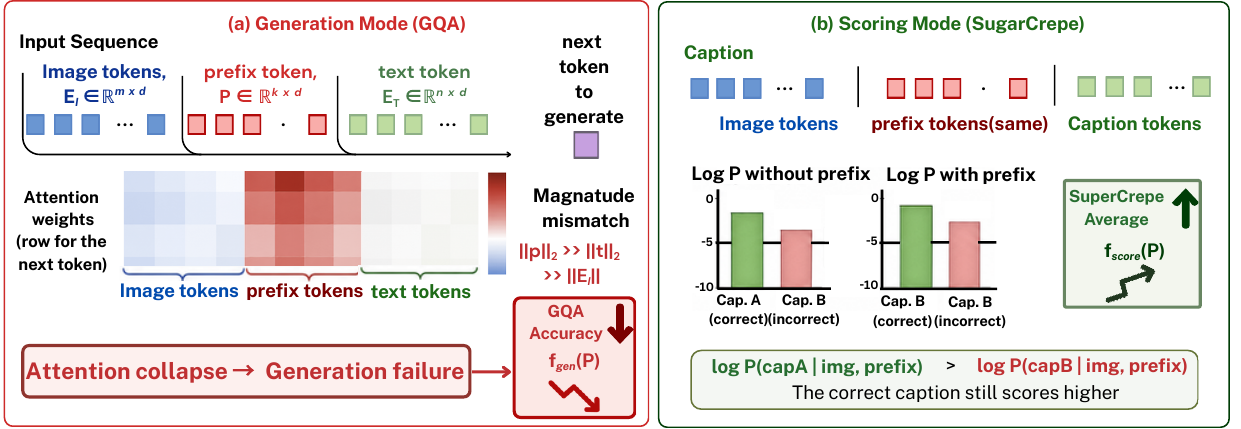}
    \caption{Why prefix tokens hurt generation but help scoring. The L2-norm mismatch distorts autoregressive attention (left) but does not affect relative log-likelihood ranking (right).}
    \label{fig:dual_contribution}
\end{figure}

\paragraph{The dual contribution of hyperbolic training.}
\label{sec:dual_contribution}
Our experiments reveal that HyperVis's relational graph serves two distinct and complementary functions (Table~\ref{tab:dual_eval}). As a \textbf{training-time regularizer}, the hyperbolic angle and entailment losses prevent LoRA adapters from overfitting to surface-level GQA patterns, yielding $+$3.82pp on GQA compared to LoRA-only training, even when prefix tokens are dropped at inference. As an \textbf{inference-time relational encoder}, the prefix tokens provide explicit compositional structure that the log-likelihood scorer can leverage, producing $+$6.25pp on SugarCrepe. These two roles are architecturally inseparable during training (the same forward pass computes both the relational losses and the prefix tokens), but their effects can be cleanly separated at inference time by toggling prefix injection (Fig.~\ref{fig:dual_contribution}).

The generation failure with prefix tokens (GQA 34.75\%) stems from a $36.5{\times}$ L2-norm mismatch between hyperbolic prefix embeddings and text embeddings, which distorts the attention distribution during autoregressive decoding. Preliminary experiments with LayerNorm and prefix dropout ($p{=}0.3$) partially close this gap (GQA recovers to 49.44\%), but fully resolving the generation--scoring tradeoff remains an important direction for future work.

\begin{table}[t]
\centering
\caption{Dual evaluation protocol (same HyperVis checkpoint). Prefix tokens hurt generative VQA but boost discriminative compositional scoring. Bold indicates the recommended protocol for each benchmark type.}
\label{tab:dual_eval}
\begin{tabular}{lcc}
\toprule
\textbf{Inference mode} & \textbf{GQA (generation)} & \textbf{SugarCrepe (scoring)} \\
\midrule
LoRA-only (no prefix)   & \textbf{61.03} & --- \\
With prefix tokens      & 34.75          & \textbf{79.94} \\
\bottomrule
\end{tabular}
\end{table}

\noindent\textbf{Top-$K$ sensitivity.}
Table~\ref{tab:ablation_topk} varies the number of prefix tokens $K$ injected into the VLM. $K{=}0$ corresponds to LoRA-only training without any relational losses. The results show two distinct regimes. For GQA (LoRA-only eval), any $K \geq 1$ recovers from the LoRA-only degradation: even $K{=}1$ achieves 60.92\%, a $+$3.71pp jump over $K{=}0$ (57.21\%), with further gains plateauing by $K{=}4$. This confirms that the presence of relational losses during training, not the number of prefix tokens at inference, drives the GQA regularization effect. For SugarCrepe, the pattern is non-monotonic: compositionality peaks at $K{=}2$ (82.34\%) then declines for $K \geq 8$ (72.81\% at $K{=}8$, 72.22\% at $K{=}16$), falling below the baseline. Too many prefix tokens introduce redundant or conflicting relational signals that degrade the log-likelihood scorer. $K{=}4$ provides the best balance (79.94\%), and we adopt it as the default.

\begin{table}[!t]
\centering
\caption{Effect of the number of prefix tokens $K$ on GQA (LoRA-only eval) and SugarCrepe (with prefix). $K{=}0$ is LoRA-only (no relational loss). Ablation models trained for 3 epochs.}
\label{tab:ablation_topk}
\begin{tabular}{lcc}
\toprule
$K$ & \textbf{GQA} & \textbf{SC Avg} \\
\midrule
0 (LoRA-only)       & 57.21 & 74.98 \\
1                   & 60.92 & 76.46 \\
2                   & 60.84 & 82.34 \\
\textbf{4 (ours)}   & \textbf{61.03} & 79.94 \\
8                   & 61.00 & 72.81 \\
16                  & 60.98 & 72.22 \\
\bottomrule
\end{tabular}
\end{table}

% =====================================================
\subsection{Discussion}
\label{sec:discussion}

\noindent\textbf{Why textual SGG injection backfires.} A discrete predicate label discards exactly the cues that make a visual relationship resolvable: the relative pose of the two regions, occlusion, and pixel-level interaction. When the LLM sees the token ``\texttt{on}'', it has lost the ability to distinguish ``standing on'', ``leaning on'', ``hovering above'', all collapsed into one symbol. Worse, SGG noise propagates symbolically: a wrong predicate is not a slightly noisy continuous feature, it is a categorically wrong word. Our continuous relation tensor preserves the spatial--visual signal end-to-end and degrades gracefully under detector noise.

\noindent\textbf{Why LoRA alone degrades GQA.} The $-$3.17pp drop from LoRA-only fine-tuning (57.21\% vs.\ 60.38\% baseline) is initially surprising: task-specific fine-tuning usually helps. We attribute this to catastrophic narrowing: LLaVA-1.5 was pretrained on diverse visual instruction data spanning many reasoning types; LoRA fine-tuning on a single dataset (GQA \texttt{train\_balanced}) narrows the model's capabilities toward surface-level GQA patterns at the expense of the broad relational reasoning needed for the testdev split. The damage concentrates on Logical ($-$7.65pp) and Compare ($-$4.41pp), the question types most dependent on multi-step reasoning. HyperVis's hyperbolic losses counteract this by forcing the LoRA weights to maintain a geometrically structured relational representation, preventing the collapse into surface-level shortcuts.

\noindent\textbf{The $\kappa{=}4.0$ revelation.} Prior hyperbolic VLMs operating on global image-text features report a curvature collapse: $\kappa$ drifts toward 0, effectively reverting the manifold to Euclidean (the \textit{curvature bottleneck} of~\cite{ramasinghe2024accept}). We observe the opposite. With our IoA-driven losses on continuous \textit{visual} features, $\kappa$ stabilises at $4.0$, an order of magnitude larger than prior reports. Continuous visual region features overlap heavily in pixel space (an arm and the body it is attached to share most of their RoI features), so the model needs the \textit{exponential} volume growth of strongly curved hyperbolic space to push relationally distinct objects apart, while spatial entailment cones simultaneously enforce that contained regions remain inside their containers' cones. High curvature is thus not a quirk; it is the geometric resolution of the tension between visual feature overlap and spatial containment.

\noindent\textbf{Euclidean vs.\ hyperbolic: a controlled decomposition.} The Euclidean ablation provides a clean decomposition of HyperVis's contributions. On GQA (LoRA-only inference), both geometries perform near-identically (60.81\% vs.\ 61.03\%), confirming that the GQA gain is a \textit{pipeline} effect. On SugarCrepe, hyperbolic geometry adds $+$4.58pp. The mechanism is the entailment cone: in Lorentz space, containment is encoded as a narrow angular cone providing geometrically precise hierarchy; in Euclidean space, this degenerates to a half-space partition, producing a ${\sim}6{\times}$ higher entailment loss at convergence. Notably, Euclidean prefix tokens damage GQA generation far less (54.06\% vs.\ 34.75\%), because there is no manifold-to-Euclidean embedding-space mismatch at injection time, corroborating the magnitude-mismatch diagnosis.

\noindent\textbf{Generalising beyond static-image VQA.} The HyperVis recipe is largely orthogonal to the choice of input modality. Replacing 2D bounding boxes with 3D spatio-temporal tubes extends the approach to video action graphs, while re-interpreting $\IoA$ as a contact/support predicate enables embodied affordance reasoning. Both inherit our finding that strongly curved space ($\kappa \gg 1$) is what makes continuous visual hierarchies usable.

\noindent\textbf{Limitations.} (i) HyperVis depends on region proposals; very small or heavily occluded objects can be missed. (ii) The $O(N^2)$ relation tensor scales quadratically; we use $N{=}36$ as a practical sweet spot. (iii) Resolving the generation--scoring tradeoff remains open (Sec.~\ref{sec:dual_contribution}). (iv) The IoA prior favours spatially compact relations; abstract relations (gaze, motion) may benefit from learned auxiliary signals.

% =====================================================
\section{Conclusion}
\label{sec:conclusion}
% =====================================================

We presented HyperVis, a framework that bypasses the SGG semantic bottleneck by routing relational reasoning over a continuous latent visual graph on the Lorentz hyperboloid. HyperVis contributes in two ways: hyperbolic relational losses regularize LoRA adapters (GQA 61.03\%, $+$3.82pp over LoRA-only), while hyperbolic prefix tokens boost compositional scoring (SugarCrepe 79.94\%, $+$6.25pp over baseline). A controlled Euclidean ablation confirms that the compositionality gain is specifically hyperbolic ($+$4.58pp), and comprehensive ablations over curvature, prefix count, and loss components decompose each contribution. The learned curvature $\kappa{=}4.0$ challenges the curvature-bottleneck narrative, demonstrating that continuous visual features require the exponential volume of strongly curved hyperbolic space.

% \section*{Acknowledgments}
% % TODO: Fill in acknowledgments

%Bibliography
\bibliographystyle{unsrt}
\bibliography{egbib}

\end{document}